\documentclass[preprint,12pt]{elsarticle}

\usepackage{amssymb}
\usepackage{amsmath}
\usepackage{float}
\usepackage{multirow}
\usepackage{adjustbox}
\usepackage{capt-of} 
\usepackage{caption}
\usepackage{graphicx,verbatim}
\usepackage{float}
\usepackage{multirow}
\usepackage{adjustbox}
\usepackage{capt-of} 
\usepackage{caption}
\usepackage{bbm}

\usepackage{booktabs}
\usepackage{algorithm}
\usepackage{algpseudocode}
\usepackage{longtable}
\usepackage{xltabular}
\usepackage{array}
\usepackage{adjustbox}
\usepackage[table]{xcolor}

\newcommand{\etal}{\textit{et al.}}

\journal{Biomedical Signal Processing and Control}

\begin{document}

\begin{frontmatter}



\title{Purification and Regulation: Comorbidity-Aware Multi-Label Few-Shot Learning for Medical Image Classification}

\author[label1]{Ying-Chih Lin}
\affiliation[label1]{organization={Department of Computer Science, National Yang Ming Chiao Tung University},
            city={Hsinchu},
            postcode={300093},
            country={Taiwan}}

\author[label2]{Po-Chih Kuo}
\affiliation[label2]{organization={Department of Computer Science, National Tsing Hua University},%Department and Organization
            city={Hsinchu},
            postcode={300093}, 
            country={Taiwan}}
            
\author[label1]{Yong-Sheng Chen\corref{cor1}}
\cortext[cor1]{Corresponding author. E-mail address: yschen@nycu.edu.tw (Y. S. Chen) }

%% Abstract
% \begin{abstract}
% %% Text of abstract
% Abstract text.
% \end{abstract}

\begin{abstract}
Multi-label few-shot learning (MLFSL) remains a significant challenge in medical image analysis (MIA).
Current metric-based meta-learning methods face two critical limitations in MIA.
First, conventional prototype generation often entangles irrelevant disease information, leading to contaminated prototypes and degraded performance.
Second, prior studies typically enforce inter-class separability in embedding space, largely neglecting the inherent correlations among diseases. 
To overcome these challenges, we propose Prototype Purification and Regulation (PPR), a novel MLFSL framework for MIA.
PPR first performs prototype purification by leveraging sample-level comorbidity scores to emphasize disease-specific features, producing purified prototypes that better characterize each disease.
Building upon these purified prototypes, PPR further addresses the underexplored problem of inter-class prototype distance in MIA by incorporating disease-level comorbidity statistics to adaptively regulate inter-class similarity, forming a comorbidity-aware embedding space.
Overall, PPR sequentially enables the model to capture pure disease features and inter-class relationships for reliable MLFSL in MIA.
Extensive experiments across four chest X-ray benchmark datasets, including cross-domain evaluation, show that PPR consistently outperforms state-of-the-art methods, significantly improving disease detection while demonstrating robust generalization and clinical applicability.

\end{abstract}

\begin{highlights}
\item This paper presents a novel multi-label few-shot learning framework for medical image analysis that jointly addresses prototype contamination and inter-class semantic issues. 
\item We introduce a prototype purification method that exploits sample-level comorbidity scores to construct purified prototypes.
\item We design a prototype regulation strategy that adaptively regulates the inter-class distance with comorbidity statistics, yielding a comorbidity-aware embedding space.
\item The designed framework achieves state-of-the-art performance on four chest X-ray datasets, including cross-domain evaluation.
    
\end{highlights}

%% Keywords
\begin{keyword}
Multi-label few-shot learning \sep Prototype purification \sep Comorbidity-aware regulation



\end{keyword}

\end{frontmatter}


    
\section{Introduction}
\label{sec:intro}

% Medical image analysis (MIA) plays a pivotal role in diagnosing a wide range of diseases and conditions in clinical practice. 
Medical image analysis (MIA) plays a pivotal role in assisting disease diagnosis and treatment planning across diverse clinical applications.
% Recently, deep learning has shown tremendous potential in MIA, significantly improving diagnostic accuracy, reducing manual workload, and accelerating the overall clinical decision process~\cite{manzari2023medvit, yue2024medmamba}. 
% Recent advances in deep learning have substantially improved diagnostic accuracy~\cite{manzari2023medvit, yue2024medmamba}.
% However, the success of these methods heavily depends on large-scale, high-quality labeled data, which is impractical in real-world healthcare scenarios.
Recent advances in deep learning have greatly improved diagnostic accuracy~\cite{manzari2023medvit,yue2024medmamba}, yet their success depends heavily on large-scale, high-quality labeled datasets that are often impractical to obtain in real-world medical settings.
To overcome this limitation, few-shot learning (FSL) has emerged as an effective paradigm for recognizing novel classes from only a few annotated samples~\cite{fei2006one,koch2015siamese,vinyals2016matching,snell2017prototypical,sung2018learning}.
Typical FSL methods adopt a meta-learning framework that trains on base classes and evaluates on unseen novel classes.
While FSL has achieved remarkable progress in single-label classification~\cite{chen2024meta,lin2024enhancing,hong2025combined,bajraktari2026few}, the extension to multi-label scenarios remains challenging, as the co-occurrence of multiple categories within an image causes feature entanglement and substantial performance degradation~\cite{simon2022meta,liu2025multi,fang2025inferring}.
This problem is particularly critical in MIA, where multiple disease patterns often coexist within a medical image~\cite{li2023patch,moukheiber2022few}.
To address this challenge, multi-label few-shot learning (MLFSL) has become an emerging research direction ~\cite{moukheiber2022few,li2023patch,liu2025multi,fang2025inferring,simon2022meta}.

\begin{figure}[t]
  \centering
  % left, down
  % \includegraphics[scale=0.48, trim = 8 10 8 0, clip]
  % \includegraphics[scale=0.145, trim = 8 10 8 0, clip]
  \includegraphics[width=\textwidth]
  {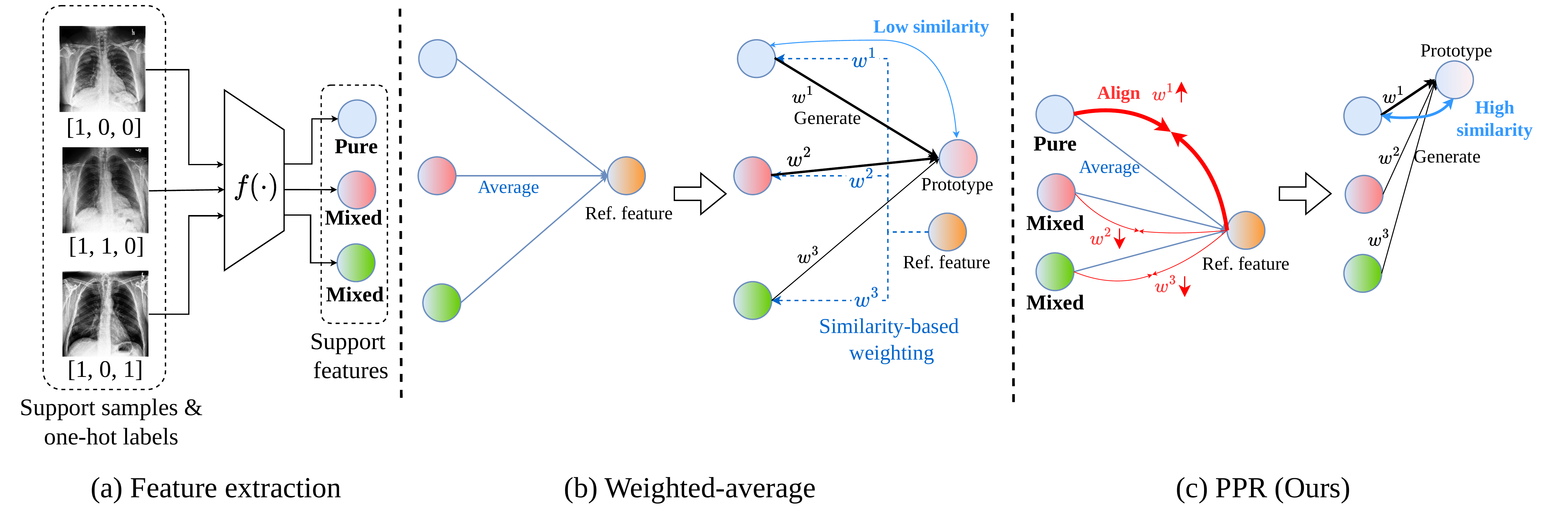}
  \caption{Comparison of prototype generation strategies in a 3-way 3-shot example. Gradient-color samples indicate mixed feature representations caused by multi-label co-occurrence, while single-colored samples correspond to pure single-disease features. (a) Support samples are encoded into the embedding space. (b) Existing methods derive a reference feature by simply averaging support features, then perform similarity-weighted averaging based on this biased reference, resulting in a contaminated prototype. (c) Our method enables the model to capture pure features by aligning the reference feature toward disease-specific representations, producing a purified prototype that better preserves disease characteristics in the embedding space.}
  \label{fig:f1}
\end{figure}

Among meta-learning paradigms, metric-based methods are widely adopted in FSL for MIA owing to their efficiency and promising performance~\cite{jin2024few,pachetti2024systematic,quinonez2025comparative,wang2026edge}, where classification is performed by measuring similarity between query features and class prototypes derived from support samples. 
However, conventional prototype construction relies on simple averaging of support features~\cite{liu2022label,yan2022inferring,simon2022meta}.
This operation inevitably entangles semantics from co-occurring diseases~\cite{liu2022label,yan2022inferring,simon2022meta}, as a single sample may contain multiple disease patterns.
To alleviate this issue, recent approaches employ similarity-weighted averaging in prototype generation~\cite{zhao2023few,an2024leveraging}.
Nevertheless, this weighting process remains biased as the reference feature for similarity weighting is typically computed from the mean of support embeddings (Fig.~\ref{fig:f1}b), which inherently contains mixed disease semantics.
Such reference features entangle irrelevant information across multiple labels, leading to misaligned similarity estimates with true class semantics.
As a result, this bias propagates noise into prototype construction, thereby limiting classification performance and reliability.
To address this, we emphasize the necessity of generating purified prototypes for reliable classification in MLFSL.

On the other hand, while prior studies have investigated the inter-class feature separation in the embedding space~\cite{liu2020negative,li2020boosting,yang2022few,jiang2022multi,song2023learning,oh2024closer}, most of them emphasize maximizing inter-class separability to improve discrimination~\cite{li2020boosting,yang2022few,jiang2022multi,song2023learning}.
In contrast, CLOSER~\cite{oh2024closer} challenges this viewpoint, showing that constraining feature dispersion by maximizing the inter-class similarity can improve novel class generalization from base classes.
While both strategies have shown promise, neither adequately accounts for the characteristics of MIA, where disease categories often exhibit partial semantic overlap due to shared pathological patterns and complex comorbidity structures.
This raises a fundamental yet underexplored question:
\textit{How should inter-class prototype distance be regulated in the embedding space to better reflect disease relationships?}
This issue is particularly important, as disease categories frequently exhibit partial semantic overlap due to shared pathological patterns and complex comorbidity structures.
This issue is particularly critical in MIA, as excessive separation may distort intrinsic disease correlations, whereas insufficient separation may cause representation ambiguity and weaken class discriminability.
Hence, we propose a prototype regulation method to maintain class discriminability while preserving intrinsic disease relationships.

In this work, we propose Prototype Purification and Regulation (PPR), a novel MLFSL framework for MIA that jointly addresses prototype contamination and the inter-class distance issues.
The key insight of purification is that samples with fewer co-occurring diseases, such as single-disease cases, provide cleaner and more disease-specific features for prototype construction.
Accordingly, we rectify the biased reference feature by aligning it toward these disease-representative embeddings, guided by the purity prior derived from sample-level comorbidity scores (Fig.~\ref{fig:f1}c).
Unlike previous methods that exploit label distribution only within the classification loss~\cite{simon2022meta}, our purification strategy operates directly in the embedding space, mitigating reference feature bias and generating discriminative purified prototypes.

In addition, we introduce a prototype regulation method that adaptively constrains inter-class distance with disease-level comorbidity statistics.
Specifically, prototypes of rarely co-occurring diseases are encouraged to be positioned farther apart than those of frequently co-occurring diseases, forming a comorbidity-aware embedding space that yields a more reasonable prototype geometry for MIA.
Consequently, diseases that share similar visual or pathological characteristics become easier to detect, as their relational context is explicitly embedded into the feature space. \\
% while preserving the generalization benefits observed in CLOSER~\cite{oh2024closer}.
\noindent The contributions of this paper are summarized as follows:
\begin{itemize}
    \item To the best of our knowledge, the proposed PPR is the first MLFSL framework for MIA that jointly addresses prototype contamination and inter-class semantic issues. 
    \item We introduce a prototype purification method that exploits sample-level comorbidity scores to construct purified prototypes.
    \item We design a prototype regulation strategy that adaptively regulates the inter-class distance with comorbidity statistics, yielding a comorbidity-aware embedding space.
    \item Comprehensive experiments on four chest X-ray datasets, including cross-domain evaluation, demonstrate that PPR achieves state-of-the-art (SoTA) performance and exhibits strong generalization and disease detection under limited supervision.
\end{itemize}

\section{Related work}
\subsection{Few-Shot Learning}
% todo relationnet 主流的先闡明
FSL aims to emulate human intelligence by generalizing the model to novel categories from only a few labeled samples, leveraging transferable knowledge distilled from previously seen classes.
The dominant paradigm for tackling FSL tasks builds on meta-learning~\cite{yang2023jlcsr}, which adopts episodic training to acquire task-agnostic meta-knowledge that can be rapidly adapted to unseen tasks with minimal supervision. 
Meta-learning approaches in FSL are generally categorized into three major branches: optimization-based, model-based, and metric-based methods.
Optimization-based methods aim to learn model initialization or adaptation strategies that enable rapid fine-tuning with only a few gradient updates on new tasks, as exemplified by MAML~\cite{finn2017model} and its extensions~\cite{nichol2018reptile,yoon2018bayesian,Flennerhag2020Meta-Learning}.
Model-based methods focus on designing architectures with inherent rapid adaptation capabilities, including memory-augmented or dynamic-parameter models, such as MANN~\cite{santoro2016meta} and SNAIL~\cite{mishra2017simple}.
However, optimization-based methods often incur expensive inner-loop updates, whereas model-based approaches rely on specialized architectures, both of which limit scalability and flexibility.
In contrast, metric-based methods~\cite{vinyals2016matching,snell2017prototypical,sung2018learning} provide a more efficient and interpretable framework by learning a discriminative embedding space where classification is performed through similarity comparison between prototypes and query features. 
Their efficiency and promising performance in MIA~\cite{jin2024few,pachetti2024systematic,quinonez2025comparative} align our work most closely with this paradigm.

% First, optimization-based methods focus on learning initialization or adaptation strategies that allow models to be fine-tuned with only a few gradient updates on new tasks, as exemplified by MAML~\cite{finn2017model} and its extensions~\cite{nichol2018reptile}. 
% Second, metric-based methods~\cite{vinyals2016matching,snell2017prototypical,sung2018learning} emphasize learning a discriminative embedding space where classification is performed through similarity comparison between prototypes and query features. 
% Representative works include Matching Network~\cite{vinyals2016matching}, Prototypical Network~\cite{snell2017prototypical}, and Relation Network~\cite{sung2018learning}. 
% Third, model-based methods center on designing architectures with inherent rapid adaptation capabilities, such as memory-augmented or dynamic-parameter models, including MANN~\cite{santoro2016meta} and SNAIL~\cite{mishra2017simple}. 
% Compared to other methods, metric-based methods circumvent the need for costly inner‑loop adaptation procedures in optimization-based approaches or specialized architectural designs in model-based techniques, maintaining computational efficiency and achieving superior performance in FSL tasks for MIA~\cite{quinonez2025comparative,pachetti2024systematic}.
% Our work is therefore most closely aligned with this line of research.
% % todo metric-based summary 加強為什麼要
% % 這領域一開始怎麼做，遇到什麼問題，然後....，遇到什麼問題，所以

\subsection{Multi-Label Few-Shot Learning}
Recent efforts have extended FSL to the multi-label setting~\cite{alfassy2019laso,chen2020knowledge,yan2022inferring,moukheiber2022few,an2024leveraging,fang2025inferring}, where each sample may be associated with multiple categories.
LaSO~\cite{alfassy2019laso} introduces label-set operation networks to model label dependencies such as union and intersection, enabling generalization to novel label combinations for MLFSL in natural image classification.
For optimization-based approaches, Moukheiber \etal~\cite{moukheiber2022few} address the recognition of rare cardiothoracic diseases by constructing a geometric ensemble of Voronoi diagrams derived from diverse multi-label training schemes, thereby facilitating effective transfer from common to rare conditions.
Aimen \etal~\cite{aimen2025generalized}  propose a Generalized Episodic Training strategy to jointly address cross-domain transfer and MLFSL.
However, the overlap between training and evaluation classes adopted in their setting is uncommon in real-world clinical scenarios.

On the other hand, for metric-based methods, BCR~\cite{an2024leveraging} improves prototype construction by jointly mining the information from instance-to-label and label-to-instance correlations with adaptive weighting, yielding notable performance improvements.
Fang \etal~\cite{fang2025inferring} propose a partial aggregation strategy that leverages multimodal information to selectively aggregate class-relevant regions for prototype estimation, thereby generating more discriminative prototypes.
Despite their effectiveness, prototype construction in MLFSL remains challenging because co-occurring diseases often introduce entangled semantics into feature representations~\cite{simon2022meta,zhao2023few,an2024leveraging}.
Existing methods typically address this issue through auxiliary modalities~\cite{fang2025inferring} or introducing additional modules~\cite{an2024leveraging}, leading to increased model complexity and limited generalization.
In contrast, we explicitly rectify the reference feature used for similarity-weighted aggregation, generating cleaner prototypes by emphasizing disease-representative features during prototype construction.

% In contrast, we address prototype contamination from a different perspective by explicitly rectifying the reference feature used for similarity weighting.
% By emphasizing disease-representative features during prototype construction, the proposed framework generates cleaner prototypes and improves multi-label few-shot recognition in medical image analysis.

% emphasizing disease-representative features, and constructing purified prototypes

% providing a simpler yet effective solution to MLFSL for MIA.

\subsection{Feature Separation} 
Prior studies have explored inter-class separation in FSL and few-shot class-incremental learning~\cite{liu2020negative,li2020boosting,hersche2022constrained,song2023learning,oh2024closer,jiang2022multi}. 
Conventional approaches generally advocate for maximizing inter-class distances in the feature space to improve class discrimination~\cite{li2020boosting,hersche2022constrained,yang2023neural,song2023learning,jiang2022multi}.
For instance, Song \etal~\cite{song2023learning} propose a semantic-aware virtual contrastive model that explicitly enlarges class margins during training to enhance feature separability.
In contrast, CLOSER~\cite{oh2024closer} challenges this intuition, arguing that constraining feature dispersion by maximizing the inter-class similarity can improve novel-class transferability from base class, as excessive class separation may hinder effective feature sharing between classes.
These conflicting findings suggest that neither global attraction nor repulsion is universally optimal, motivating the need for adaptive inter-class regulation.

\section{Method}
\label{sec:method}
\subsection{Preliminaries}
Following standard practice~\cite{simon2022meta}, a given dataset is partitioned into three disjoint disease subsets: a base set \(C_{\mathrm {base}}\), a validation set \(C_{\mathrm{val}}\), and a novel set \(C_{\mathrm{novel}}\). 
The partitioning is conducted in descending order of class samples, ensuring that low-prevalence or rare diseases are assigned to the novel set \(C_{\mathrm{novel}}\) to better reflect clinical deployment scenarios~\cite{ouahab2022self,wen2025few}. Patient identities are kept mutually exclusive across all subsets to prevent data leakage.

% Technically, episodic meta-training can be formulated as an \(N\)-way \(K\)-shot problem~\cite{simon2022meta,an2024leveraging}. 
Technically, episodic meta-training is formulated as an \(N\)-way \(K\)-shot problem~\cite{simon2022meta,an2024leveraging}, where each episode $\mathcal{E}$ samples a set of disease classes $C_{\mathcal{E}}$, with $|C_{\mathcal{E}}| = N$.
An episode $\mathcal{E}$ consists of a support set \(\mathcal {S} = \{ (\textbf{x}^i, \textbf{y}^i)\}^{|\mathcal{S}|}_{i = 1}\) and a query set \(\mathcal {Q} = \{ (\textbf{x}^i, \textbf{y}^i)\}^{|\mathcal{Q}|}_{i = 1}\), where an image \(\textbf{x}\) is associated with a multi-label vector $\textbf{y} = [ y_1,\dots, y_{N} ] \in \{0,1\}^{N}$ and $y_c = 1$ indicates the presence of the $c$-th class.
To accommodate multi-label settings with varying class sample counts per episode, we enforce a minimum of \(K\) samples per class in each episode, ensuring sufficient data for meta-training and meta-testing.

% Each episode $\mathcal{E}$ consists of a support set \(\mathcal {S} = \{ (\textbf{x}^i, \textbf{y}^i)\}^{|\mathcal{S}|}_{i = 1}\) and a query set \(\mathcal {Q} = \{ (\textbf{x}^i, \textbf{y}^i)\}^{|\mathcal{Q}|}_{i = 1}\), where an image \(\textbf{x}\) is associated with a multi-label vector $\textbf{y} = [ y_1,\dots, y_{N} ] \in \{0,1\}^{N}$ and $y_c = 1$ indicates the presence of the $c$-th class.
% The set of sample diseases per episode is represented by $C_{\mathcal{E}}$, with $|C_{\mathcal{E}}| = N$.
% In MLFSL, each image can contain multiple co-occurring disease labels, leading to variable numbers of positive samples per class across episodes.
% To ensure balanced sampling, each episode is constructed such that every selected class includes at least \(K\) samples, providing sufficient data for both meta-training and meta-testing.

\begin{figure*}[t]
  \centering
  \includegraphics[width=\textwidth]
  {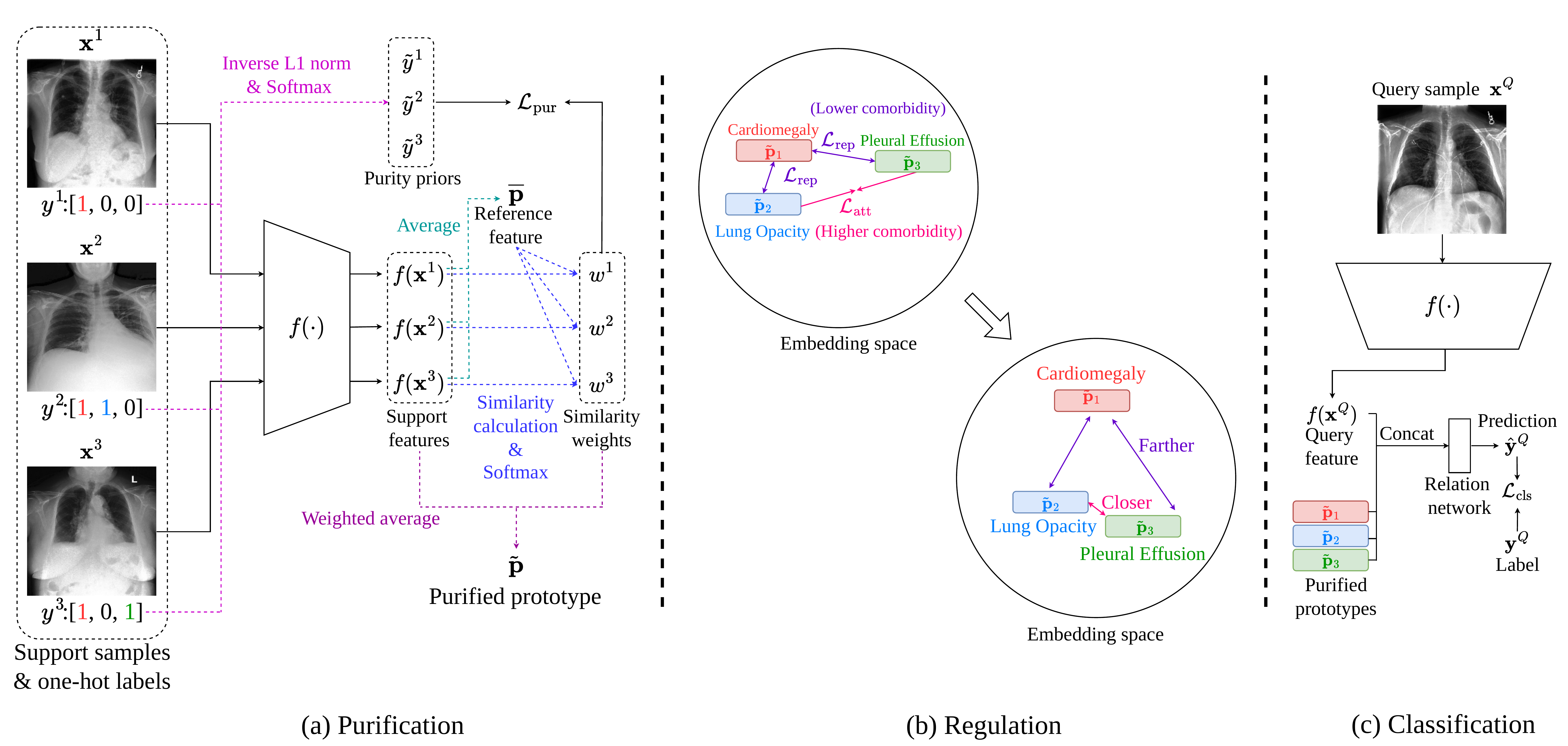}
  \caption{Overview of the proposed PPR framework in a 3-way 3-shot example. (a) Purification: The reference feature is aligned toward purer representations by leveraging the label distribution derived from the sample comorbidity score as a purity prior, thereby emphasizing discriminative features and generating purified prototypes.
  (b) Regulation: prototype similarities are adaptively constrained based on disease comorbidity statistics to capture inter-disease relationships. (c) Classification: query features are concatenated with the purified prototypes and fed into the relation network for final prediction.}
  \label{fig:overall}
\end{figure*}

\subsection{Overview}
As illustrated in Fig.~\ref{fig:overall}, the proposed PPR framework comprises three key components: \textbf{purification}, \textbf{regulation}, and \textbf{classification}.
The purification stage generates purified class prototypes by emphasizing disease-representative features, while the regulation stage incorporates disease relationship information to regulate inter-class prototype similarity. 
In the classification stage, we employ a relation network~\cite{sung2018learning, simon2022meta} to model the relationship between query features and the purified class prototypes for final prediction.
The relation network architecture follows the original design in~\cite{sung2018learning}, comprising two convolutional blocks.
The details of the purification and regulation stages are presented in the following sections.

% $c = 1 \dots |C_{\mathcal{E}}|$
\subsection{Prototype purification}
The conventional prototype $\overline {\bf p}_c$ for class $c$, with $c \in C_{\mathcal{E}}$, is typically obtained by the mean of support features:
\begin{equation}
    \begin{aligned} 
     \overline {\bf p}_c = \frac{\sum_{(\textbf{x},\textbf{y}) \in \mathcal{P}_c} f(\textbf{x}) }{ \left | \mathcal{P}_c\right |}
    \end{aligned} \, ,
\end{equation}
% \(\mathcal{P}_c = \begin{Bmatrix} {\textbf{x}}^i \mid ({\textbf{x}}^i, {\textbf{y}}^i) \in \mathcal{S}, \ y^i_c = 1  \end{Bmatrix}\)
where \(\mathcal{P}_c = \begin{Bmatrix} ({\textbf{x}}, {\textbf{y}}) \ | \ ({\textbf{x}}, {\textbf{y}}) \in \mathcal{S} \ , \ y_c = 1  \end{Bmatrix}\) represents the subset of support samples labeled with disease $c$. 
% with $i = 1, \dots, \left | \mathcal{S}\right |$
The notation \(f(\cdot)\) denotes the feature encoder.
In multi-label settings, however, the mean prototype $\overline {\bf p}_c$ can be contaminated by mixed semantics arising from co-occurring diseases, resulting in an impure and less discriminative class representation.
To address this issue, we first utilize $\overline {\bf p}_c$ as the reference feature and compute the cosine similarity between each support embedding and $\overline {\bf p}_c$:
\begin{equation}
    \begin{aligned} 
    {z}^i_c = \frac{f(\textbf{x}^i) \cdot \overline {\bf p}_c}{\left\|f(\textbf{x}^i)\right\|   \left\|\overline {\bf p}_c  \right\|} , \ i=1,\dots,|\mathcal{P}_c| ,(\textbf{x}^i,\textbf{y}^i)\in \mathcal{P}_c
    \end{aligned} \, .
\end{equation}
Subsequently, the normalized similarity weight of each sample $w^i_c$ is computed via a softmax function, and the purified prototype $\tilde {\bf p}_c$ is calculated as a weighted combination of support features:
\begin{equation}
    \begin{aligned} 
     \tilde {\bf p}_c = \sum_{i=1}^{|\mathcal{P}_c|} w^i_c  \ f(\textbf{x}^i)
     , \ w^i_c = \frac{\mathrm{exp}({ z}^i_c /  \tau)}{\sum_{j=1}^{|\mathcal{P}_c|}\mathrm{exp}({ z}^j_c   /\tau)}
     \end{aligned} \, ,
\end{equation}
% \begin{equation}
%     \begin{aligned} 
%       w^i_c = \frac{\mathrm{exp}({ z}^i_c /  \tau)}{\sum_{j=1}^{|\mathcal{P}_c|}\mathrm{exp}({ z}^j_c   /\tau)}
%     \end{aligned} \, ,
% \end{equation}
where $\tau$ is a trainable temperature coefficient.
% Accordingly, the purified prototype $\tilde {\bf p}_c$ is calculated as a weighted combination of support features:
% \begin{equation}
%     \begin{aligned} 
%      \tilde {\bf p}_c = \sum_{i=1}^{|\mathcal{P}_c|} w^i_c \cdot f(\textbf{x}^i)
%     \end{aligned} \, .
% \end{equation}
However, relying solely on similarity-based reweighting is insufficient to mitigate prototype contamination, as it still depends on a biased reference feature $\overline {\bf p}_c$ derived from entangled disease semantics.
To rectify this bias, we explicitly guide the model toward learning from disease-specific clean representations and design the purification loss $\mathcal{L}_{\text{pur}}$.
The key intuition of this strategy is that samples with fewer co-occurring diseases, corresponding to a lower comorbidity score, are more likely to exhibit purer and more discriminative disease features.
Therefore, we define a purity prior ${\tilde y}^i_c$ for each sample within an episode based on its comorbidity score as:
\begin{equation}
    \begin{aligned} 
     {\tilde y}^i_c = \frac{\mathrm{exp}{(\left\| {\bf y}^i\right\|_1^{-1})}} {\sum_{j=1}^{|\mathcal{P}_c|} \mathrm{exp}{(\left\| {\bf y}^{j}\right\|_1^{-1})} }
    \end{aligned} \, .
\end{equation}
Here, the purity prior ${\tilde y}^i_c$ assigns higher values to samples with lower comorbidity situations within each episode.
The purification loss is then formulated as the KL divergence between the purity prior ${\tilde y}^i_c$ and the similarity weight $w^i_c$:
\begin{equation}
    \begin{aligned} 
     \mathcal{L}_{\text{pur}} = \sum_{c \in C_{\mathcal{E}}}\frac{1}{|\mathcal{P}_c|}\sum_{  i=1 }^{|\mathcal{P}_c|}{ \tilde y}^i_c \text{log} \frac{ { \tilde y}^i_c}{w^i_c} + (1-{ \tilde y}^i_c)\text{log}\frac{1-{ \tilde y}^i_c}{1-w^i_c} \ 
    \end{aligned} \, ,
\end{equation}
By aligning similarity weights with these priors, the model is explicitly guided to learn from purer disease features, while implicitly aligning the reference feature toward cleaner, disease-specific representations in the embedding space.
This increases the similarity weights between the reference and disease-representative features, producing purified prototypes and improving classification reliability in MLFSL.
Unlike existing prototype contamination mitigation methods~\cite{zhao2023few,an2024leveraging,fang2025inferring} that reweight features based on a potentially biased reference, our purification strategy directly targets the biased reference feature, fundamentally addressing the mixed-semantic issue in multi-label settings.

% The proposed purification strategy enables the model to capture pure disease semantics in the embedding space, allowing for generalization to unseen diseases during inference without requiring the purity prior information.

\subsection{Prototype regulation}
In MLFSL for MIA, enforcing excessive separation between prototypes might be problematic, as many diseases co-occur and share correlated feature patterns.
To account for such intrinsic inter-disease relationships, we leverage comorbidity information to adaptively constrain inter-class similarity.
Specifically, we first quantify the comorbidity statistics $j$ between diseases $m$ and $n$ using the Jaccard index, defined within the range [0,1]:
\begin{equation}
    \begin{aligned} 
    j_{mn} = D_{mn} / (D_m + D_n-D_{mn})
    \end{aligned} \, ,
\end{equation}
where $D_m$ and $D_n$ denote the sample counts of class $m$ and $n$, and $D_{mn}$ is the number of samples in which both categories co-occur in the training dataset.
Subsequently, we introduce the prototype regulation loss $\mathcal{L}_{\mathrm{reg}}$, which can be decomposed into two complementary components: 
\begin{equation}
\begin{aligned}
\mathcal{L}_{\text{reg}} = \mathcal{L}_{\text{att}} + \mathcal{L}_{\text{rep}} \ .
\end{aligned}
\end{equation}
For co-occurring disease pairs that are overly distant in the embedding space under the comorbidity criterion $j_{mn}$, we facilitate feature proximity through the attraction loss $\mathcal{L}_{\text{att}}$:
% (i.e., $s_{mn}<j_{mn}$)
% \begin{equation}
%     \begin{aligned}
% \mathcal{L}_{\text{att}} &= \sum_{\substack{m,n\in C_{\mathcal{E}} \\ m\neq n, \  \ j_{mn} > 0}} \max\left(0, j_{mn} - s_{mn} \right)  \, ,
%     \end{aligned}
% \end{equation}
\begin{equation}
    \begin{aligned}
\mathcal{L}_{\text{att}} &= \sum_{\substack{m,n\in C_{\mathcal{E}} \\ m\neq n, \  \ s_{mn}<j_{mn}}} j_{mn} - s_{mn}   \, ,
    \end{aligned}
\end{equation}
where $s_{mn}$ represents the cosine similarity between the purified prototypes $\tilde {\bf p}_m$ and $\tilde {\bf p}_n$.
Conversely, for disease pairs that are undesirably close in the embedding space, we impose a repulsion loss $\mathcal{L}_{\text{rep}}$ to enhance feature dissimilarity:
\begin{equation}
\mathcal{L}_{\text{rep}} = \sum_{\substack{m,n\in C_{\mathcal{E}} \\ m\neq n}}
\begin{cases}
\max\left(0, s_{mn} - \delta_n \right) & \text{if } j_{mn} = 0 \\
\max\left(0,s_{mn} - (j_{mn} + \delta_p )\right) & \text{otherwise} \, ,
\end{cases}
\end{equation}
where $\delta_n$ and $\delta_p$ are negative and positive margin hyperparameters~\cite{li2020boosting,liu2020negative}, respectively.
Since the MIA datasets commonly include a \textit{No Finding} category representing healthy cases, we introduce a specific condition in the loss function where $j_{mn} = 0$, as \textit{No Finding} has no comorbidity correlations with other diseases.
To ensure that the \textit{No Finding} class remains sufficiently distinct from other disease classes, a negative margin $\delta_n$ is applied to further enlarge the inter-class semantic margin.
In contrast, a positive margin $\delta_p$ is introduced to maintain a mild similarity constraint among co-occurring diseases and prevent over-separation.
% For non-co-occurring diseases (e.g., No Finding with others), a negative margin $\delta_n$ enforces a minimum dissimilarity constraint, whereas for co-occurring diseases, a positive $\delta_p$ maintains a mild similarity margin and prevents overfitting to dataset-specific comorbidity statistics.
% Finally, the regulation loss $\mathcal{L}_{\text{reg}}$ can be defined as:
% \begin{equation}
% \begin{aligned}
% \mathcal{L}_{\text{reg}} = \mathcal{L}_{\text{att}} + \mathcal{L}_{\text{rep}}
% \end{aligned}
% \end{equation}
In this manner, inter-class similarity is adaptively regulated based on $j_{mn}$ rather than applying a uniform constraint~\cite{yang2022few,oh2024closer}.
This design preserves the generalization benefit of promoting inter-class proximity as in CLOSER~\cite{oh2024closer} while accounting for disease relationships, forming a comorbidity-aware embedding space.

% , and improving disease detection by leveraging shared pathological patterns and comorbidity correlations.

% This adaptive regulation mechanism effectively balances prototype compactness and separability, allowing the learned embedding space to reflect clinically meaningful disease relationships and improve detection by leveraging shared pathological patterns and comorbidity correlation.

\subsection{Overall loss function and inference}
Overall, PPR exploits comorbidity information to purify class prototypes and adaptively regulate inter-class prototype distance for reliable MLFSL in MIA, as defined by the following objective:
% The overall optimization objective is formulated as follows:
\begin{equation}
    \begin{aligned} 
    \mathcal{L}_{\text{all}} = \mathcal{L}_{\text{cls}} + \lambda_1\mathcal{L}_{\text{pur}} + \lambda_2\mathcal{L}_{\text{reg}}
    \end{aligned} \, ,
\end{equation}
where the classification loss $\mathcal{L}_{\text{cls}}$ can be defined as:
\begin{equation}
    \begin{aligned} 
    \mathcal{L}_{\text{cls}} = -\frac{1}{|\mathcal{Q}|}\sum_{i}^{|\mathcal{Q}|}\sum_{c \in C_{\mathcal{E}}}y^i_c \ \text{log}\ \hat{y}^i_c
    \end{aligned} \, .
\end{equation}
The notation $\hat{y}^i_c$ represents the prediction output for the class $c$ of $i$-th query sample from the relation network. 
The trade-off hyperparameters $\lambda_1$ and $\lambda_2$ balance the contribution of the prototype purification and regulation objectives, respectively.
To ensure a stable optimization process, $\lambda_2$ is gradually increased through a linear warm-up schedule based on the current epoch, allowing the model to construct purified prototypes before applying inter-class regulation.
Note that the model does not require the purity prior or co-occurrence statistics during inference.
% In the inference stage, the model operates without requiring the purity prior or additional fine-tuning.
The purified prototypes constructed from the support samples of novel classes are concatenated with the query features and fed into the relation network to produce the final predictions.

\section{Results}
\subsection{Datasets and preprocessing}
We evaluate PPR on four widely used multi-label chest X-ray (CXR) datasets, as CXR analysis represents one of the most prevalent and clinically significant multi-label challenges in MIA.
To ensure a fair and consistent few-shot evaluation, we introduce new dataset splits. 
Note that the most prevalent diseases form the base set for meta-training, while moderately frequent and rare diseases constitute the validation and novel test sets, respectively. \\
\noindent \textbf{CheXpert}~\cite{irvin2019chexpert}. The CheXpert dataset contains 224,316 chest radiographs from 65,240 patients with 14 annotated observations extracted from radiology reports. 
Following prior works~\cite{saporta2022benchmarking,hou2024icon}, we exclude \textit{Support Devices} and \textit{Pleural Other}.
The base classes include \textit{No Finding}, \textit{Cardiomegaly}, \textit{Lung Opacity}, \textit{Edema}, \textit{Atelectasis}, and \textit{Pleural Effusion}.
The validation classes are \textit{Enlarged Cardiomediastinum}, \textit{Consolidation}, and \textit{Pneumothorax}, while the novel classes are \textit{Lung Lesion}, \textit{Pneumonia}, and \textit{Fracture}. \\
\noindent \textbf{NIH ChestXRay14} (CXR14)~\cite{wang2017chestx}.
The CXR14 dataset includes 112,120 frontal images across 14 disease classes.
The base classes are \textit{No Finding}, \textit{Infiltration}, \textit{Effusion}, \textit{Atelectasis}, \textit{Nodule}, \textit{Mass}, \textit{Pneumothorax}, and \textit{Consolidation}.
The validation classes are \textit{Pleural Thickening}, \textit{Cardiomegaly}, and \textit{Emphysema}, and the novel classes include \textit{Edema}, \textit{Fibrosis}, \textit{Pneumonia}, and \textit{Hernia}. \\
\noindent \textbf{MIMIC-CXR} (MIMIC)~\cite{johnson2019mimic}.
The MIMIC dataset contains 377,110 images with 14 labels, we exclude \textit{Support Devices} and \textit{Pleural Other}.
The base classes comprise \textit{No Finding}, \textit{Cardiomegaly}, \textit{Lung Opacity}, \textit{Edema}, \textit{Atelectasis}, and \textit{Pleural Effusion}.
The validation set includes \textit{Pneumonia}, \textit{Consolidation}, and \textit{Pneumothorax}, and the novel set contains \textit{Lung Lesion}, \textit{Enlarged Cardiomediastinum}, and \textit{Fracture}. \\
\noindent \textbf{CXR-LT 2024}~\cite{holste2023cxr}. The CXR-LT 2024 dataset extends MIMIC-CXR with additional annotations for rare diseases, enriching the label diversity.
In this study, we focus on Task 3 of the challenge, which assesses generalization to unseen disease categories, including \textit{Bulla}, \textit{Cardiomyopathy}, \textit{Hilum Enlargement}, \textit{Osteopenia}, and \textit{Scoliosis}. \\
\noindent \textbf{Data Preprocessing}. 
% In all experiments, we focus on the widely used frontal-view chest X-ray (CXR) images in the clinical scenario~\cite{lin2025weighted}.
All experiments are conducted on frontal-view chest X-ray (CXR) images commonly used in clinical diagnostics.
The images are resized to 224 $\times$ 224 and data augmentations include random horizontal flipping and random rotation to enhance robustness.

\subsection{Implementation details}
The entire framework is trained end-to-end for 1k epochs using the Adam optimizer with an initial learning rate of \(5 \times 10^{-4}\), managed by a cosine learning rate scheduler. 
The episode is set to 200, 250, and 1k per epoch for training, validation, and testing, respectively~\cite{simon2022meta}.
ResNet-50~\cite{he2016deep} is adopted as the backbone following prior multi-label MIA studies~\cite{huang2024label,lin2025weighted,aimen2025generalized}, and the best validation checkpoint is used for testing.
% In alignment with previous multi-label medical image classification studies~\cite{huang2024label, lin2025weighted,aimen2025generalized}, ResNet-50~\cite{he2016deep} is adopted as our backbone network. 
% During meta-testing, the model that achieves the best validation performance is utilized for inference.
For loss balancing, since $\lambda_1$ and $\lambda_2$ control the relative contributions of $\mathcal{L}_{\text{pur}}$ and $\mathcal{L}_{\text{reg}}$, which operate on different numerical scales. 
We set ($\lambda_1$, $\lambda_2$) to (2.0, 0.01), (3.5, 0.01), and (7.5, 0.008) for CheXpert~\cite{irvin2019chexpert}, MIMIC~\cite{johnson2019mimic}, and CXR14~\cite{wang2017chestx}, respectively.
The margin hyperparameters are empirically fixed at $\delta_p = 0.05$ and $\delta_n = -0.1$ for all datasets to ensure robustness.
All experiments are implemented in PyTorch and executed on an NVIDIA RTX 4090 GPU.

\subsection{Performance comparison}
\subsubsection{Overall performance comparison}
We conduct a comprehensive comparison between PPR and recent SoTA MLFSL approaches that follow the same episodic training paradigm, including ProtoNet+NLC~\cite{simon2022meta}, RelationNet~\cite{simon2022meta}, LPN+NLC~\cite{simon2022meta}, and BCR~\cite{an2024leveraging}.
Performance is evaluated using seven key metrics: macro/ micro F1, macro/ micro recall, macro/ micro precision, and mAP.
To ensure a fair comparison, all methods are trained with the same backbone and experimental settings.
All the source codes are obtained from official repositories.

\begin{table*}[t]
\centering
\caption{Results of overall performance comparison (\%) in the 3-way 5-shot setting. The best scores are highlighted in bold and the second-best in blue.}
\label{tab:overall5}
\resizebox{\textwidth}{!}{
\begin{tabular}{c|lccccccc}
\hline
                           & \multicolumn{1}{c}{}                         & \multicolumn{7}{c}{5-shot}                 \\ \cline{3-9} 
\multirow{-2}{*}{Dataset}  & \multicolumn{1}{c}{\multirow{-2}{*}{Method}} & maF1                         & miF1                                  & maR                          & miR                          & maP                          & miP                          & mAP  \\ \hline
                           & ProtoNet+NLC~\cite{simon2022meta}                                 & 34.75                        & 37.13                                 & 36.73                        & 36.73                        & \textbf{38.75}               & \textbf{38.73}               & 39.54  \\
                           & RelationNet~\cite{simon2022meta}                                  & 41.24                        & 44.11                                 & 60.22                        & 60.23                        & 36.03                        & 36.02                        & {\color[HTML]{0000FF} 46.46} \\
                           & LPN+NLC~\cite{simon2022meta}                                      & 32.17                        & 37.06                                 & 35.60                         & 35.60                         & {\color[HTML]{0000FF}37.52}                       & {\color[HTML]{0000FF}37.54}                      & 39.45                        \\
                           & BCR~\cite{an2024leveraging}                                          & {\color[HTML]{0000FF} 46.12} & {\color[HTML]{0000FF} 46.78}          & {\color[HTML]{0000FF} 70.38} & {\color[HTML]{0000FF} 70.38} & 35.51                        & 35.51                        & \textbf{46.62}                \\
\multirow{-5}{*}{CheXpert~\cite{irvin2019chexpert}} & PPR (Ours)                                  & \textbf{49.82}               & \textbf{50.41}                        & \textbf{86.84}               & \textbf{86.85}               &  35.72 &  35.72 & 46.37                        \\ \hline
                           & ProtoNet+NLC~\cite{simon2022meta}                                 & 34.19                        & 36.58                                 & 35.66                        & 35.66                        & \textbf{37.64}               & \textbf{37.64}               & 39.28                         \\
                           & RelationNet~\cite{simon2022meta}                                  & 27.34                        & 31.61                                 & 31.69                        & 31.70                         & 36.09                        & 36.09                        & \textbf{46.95}                \\
                           & LPN+NLC~\cite{simon2022meta}                                      & 31.86                        & 35.35                                 & 34.56                        & 34.56                        & {\color[HTML]{0000FF} 36.47} & {\color[HTML]{0000FF} 36.48} & 38.74                         \\
                           & BCR~\cite{an2024leveraging}                                          & {\color[HTML]{0000FF} 42.66} & {\color[HTML]{0000FF} 43.43}          & {\color[HTML]{0000FF} 57.91} & {\color[HTML]{0000FF} 57.91} & 35.55                        & 35.55                        & 46.42                        \\
\multirow{-5}{*}{MIMIC~\cite{johnson2019mimic}}    & PPR (Ours)                                  & \textbf{46.27}               & \textbf{47.08}                        & \textbf{70.87}               & \textbf{70.88}               & 35.54                        & 35.55                        & {\color[HTML]{0000FF} 46.67}  \\ \hline
                           & ProtoNet+NLC~\cite{simon2022meta}                                 & 40.76                        & 42.74                                 & 41.49                        & 41.49                        & 43.93               & 43.91               & 42.37                         \\
                           & RelationNet~\cite{simon2022meta}                                  & 33.82                        & 35.48                                 & 31.50                        & 31.50                        & 41.91                        & 41.89                        & 51.57               \\
                           & LPN+NLC~\cite{simon2022meta}                                      & {\color[HTML]{0000FF} 42.17} & {\color[HTML]{0000FF} 44.79} & {\color[HTML]{0000FF} 43.04} & {\color[HTML]{0000FF} 43.04} & {\color[HTML]{0000FF} 45.57} & {\color[HTML]{0000FF} 45.55} & 43.88                         \\
                           & BCR~\cite{an2024leveraging}                                          & 33.27                        & 35.53                                 & 26.65                        & 26.66                        & \textbf{57.08}               & \textbf{57.19}               & \textbf{60.81}               \\
\multirow{-5}{*}{CXR14~\cite{wang2017chestx}}    & PPR (Ours)                                  & \textbf{43.93}               & \textbf{44.92}                        & \textbf{53.78}               & \textbf{53.78}               & 40.12                        & 40.10                         & {\color[HTML]{0000FF} 51.64}  \\ \hline
\end{tabular}
}
\end{table*}

\begin{table*}[t]
\centering
\caption{Results of overall performance comparison (\%) in the 3-way 10-shot setting. The best scores are highlighted in bold and the second-best in blue.}
\label{tab:overall10}
\resizebox{\textwidth}{!}{
\begin{tabular}{c|lccccccc}
\hline
                           & \multicolumn{1}{c}{}                                                   & \multicolumn{7}{c}{10-shot}                                                             \\ \cline{3-9} 
\multirow{-2}{*}{Dataset}  & \multicolumn{1}{c}{\multirow{-2}{*}{Method}} & maF1                         & miF1                         & maR                          & miR                          & maP                          & miP                          & mAP                          \\ \hline
                           & ProtoNet+NLC~\cite{simon2022meta}                                                       & 36.69                        & 38.81                        & 38.48                        & 38.47                        & {\color[HTML]{0000FF} 40.87} & {\color[HTML]{0000FF} 40.82} & 40.30                        \\
                           & RelationNet~\cite{simon2022meta}                                  & {\color[HTML]{0000FF} 48.34} & {\color[HTML]{0000FF} 49.09} & {\color[HTML]{0000FF} 79.87} & {\color[HTML]{0000FF} 79.87} & 35.69                        & 35.69                        & {\color[HTML]{0000FF} 46.40} \\
                           & LPN+NLC~\cite{simon2022meta}                                                             & 44.73                        & 47.07                        & 47.38                        & 47.45                        & \textbf{47.08}               & \textbf{47.13}               & 44.60                        \\
                           & BCR~\cite{an2024leveraging}                                                         & 47.68                        & 47.85                        & 74.80                        & 74.80                         & 35.25                        & 35.25                        & 46.21                        \\
\multirow{-5}{*}{CheXpert~\cite{irvin2019chexpert}} & PPR (Ours)                                                          & \textbf{50.14}               & \textbf{50.35}               & \textbf{87.62}               & \textbf{87.63}               & 35.59                        & 35.59                        & \textbf{46.53}               \\ \hline
                           & ProtoNet+NLC~\cite{simon2022meta}                                                       & 34.73                        & 36.84                        & 35.86                        & 35.86                        & \textbf{37.83}               & \textbf{37.82}               & 39.40                        \\
                           & RelationNet~\cite{simon2022meta}                                                 & 45.22                        & 45.49                        & 65.21                        & 65.21                        & 35.28                        & 35.27                        & 45.98                        \\
                           & LPN+NLC~\cite{simon2022meta}                                                              & 33.13                        & 36.70                        & 36.21                        & 36.21                        & {\color[HTML]{0000FF} 37.76} & {\color[HTML]{0000FF} 37.74} & 39.12                        \\
                           & BCR~\cite{an2024leveraging}                                                                 & {\color[HTML]{0000FF} 45.79} & {\color[HTML]{0000FF} 45.98} & {\color[HTML]{0000FF} 66.48} & {\color[HTML]{0000FF} 66.48} & 35.54                        & 35.54                        & \textbf{46.55}               \\
\multirow{-5}{*}{MIMIC~\cite{johnson2019mimic}}    & PPR (Ours)                                   & \textbf{45.94}               & \textbf{46.84}               & \textbf{71.40}               & \textbf{71.40}               & 35.26                        & 35.26                        & {\color[HTML]{0000FF} 46.09} \\ \hline
                           & ProtoNet+NLC~\cite{simon2022meta}                                                         & {\color[HTML]{0000FF} 44.00} & 45.70                        & {\color[HTML]{0000FF} 44.46} & {\color[HTML]{0000FF} 44.46} & {\color[HTML]{0000FF} 47.04} & {\color[HTML]{0000FF} 47.03} & 44.30                        \\
                           & RelationNet~\cite{simon2022meta}                                                 & 38.03                        & 39.59                        & 40.40                        & 40.40                        & 39.94                        & 39.96                        & 50.44                        \\
                           & LPN+NLC~\cite{simon2022meta}                                                              & 43.16                        & {\color[HTML]{0000FF} 45.94} & 44.45                        & 44.45                        & 46.93                        & 46.94                        & 44.28                        \\
                           & BCR~\cite{an2024leveraging}                                                         & 41.27                        & 42.63                        & 39.56                        & 39.56                        & \textbf{48.65}               & \textbf{48.65}               & \textbf{56.21}               \\
\multirow{-5}{*}{CXR14~\cite{wang2017chestx}}    & PPR (Ours)                                   & \textbf{47.15}               & \textbf{47.85}               & \textbf{56.05}               & \textbf{56.05}               & 42.79                        & 42.79                        & {\color[HTML]{0000FF} 54.21} \\ \hline
\end{tabular}
}
\end{table*}

As shown in Table~\ref{tab:overall5} and Table~\ref{tab:overall10}, PPR consistently achieves SoTA results across all three datasets in both 3-way 5-shot and 3-way 10-shot settings, attaining the highest F1 score and demonstrating superior generalization to unseen diseases.
While recall and precision trade off naturally, recall is particularly critical in MIA, as missed disease cases pose a much higher risk than false alarms.
% Notably, PPR delivers substantial gains in macro-recall, a crucial metric for minimizing false negatives that carry higher clinical risk than false positives.
% a crucial metric in MIA, where timely and accurate disease detection is vital. 
% PPR delivers notable gains in macro-recall, highlighting its effectiveness in improving disease detection.
PPR surpasses the second-best method in macro-recall by 16.46\% and 7.75\% on CheXpert~\cite{irvin2019chexpert}, by 12.96\% and 4.92\% on MIMIC~\cite{johnson2019mimic}, and by 10.74\% and 11.59\% on CXR14~\cite{wang2017chestx} in the 5-shot and 10-shot settings, respectively.
% PPR surpasses the second-best method in macro-recall by 16.46\% in the 5-shot setting and 7.75\% in the 10-shot setting on CheXpert~\cite{irvin2019chexpert}, 12.96\% in the 5-shot setting and 4.92\%  in the 10-shot setting on MIMIC~\cite{johnson2019mimic}, and 10.74\% in the 5-shot setting and 11.59\% in the 10-shot setting on CXR14~\cite{wang2017chestx}.
The results highlight that PPR maintains robust performance even under more challenging few-shot conditions, effectively enhancing novel disease detection with limited supervision.
Notably, these improvements indicate that PPR effectively reduces false negatives while maintaining competitive precision, a crucial requirement for clinical applicability.

% Overall, PPR not only enhances the discriminative power of prototypes but also advances the frontier of few-shot MIA toward more clinically reliable recognition.

% In contrast to the other methods that need additional modules for capturing the label-related information~\cite{simon2022meta,an2024leveraging}, we simply utilize two loss function designs, which are purify and separate, to enhance the ability of model to recognize novel diseases.

\begin{table*}[]
\centering
\caption{Results of individual disease performance comparison using F1 score (\%) in the 3-way 5-shot and 3-way 10-shot settings. The best scores are highlighted in bold and the second-best in blue.}
\label{tab:novel}
\resizebox{\textwidth}{!}{
\begin{tabular}{l|c|ccc|ccc|cccc}
\hline
                         &                        & \multicolumn{3}{c|}{CheXpert~\cite{irvin2019chexpert}}                                                     & \multicolumn{3}{c|}{MIMIC~\cite{johnson2019mimic}}                                                       & \multicolumn{4}{c}{CXR14~\cite{wang2017chestx}}                                                                                        \\ \cline{3-12} 
\multirow{-2}{*}{Method} & \multirow{-2}{*}{shot} & Pneumonia                    & Lung Lesion                  & Fracture                     & Enl. Card.     & Lung Lesion                  & Fracture                     & Hernia                       & Edema                        & Pneumonia                    & Fibrosis                     \\ \hline
ProtoNet+NLC~\cite{simon2022meta}             &                        & 37.16                        & 37.03                        & 37.08                        & 36.64                        & 36.57                        & 36.64                        & 42.67                        & 42.71                        & 42.69                        & 42.57                        \\
RelationNet~\cite{simon2022meta}              &                        & 45.01                        & 45.11                        & 45.10                         & 33.70                         & 33.60                         & 33.76                        & 35.95                        & 35.99                        & 35.87                        & 35.95                        \\
LPN+NLC~\cite{simon2022meta}                  &                        & 36.44                        & 36.48                        & 36.58                        & 35.46                        & 35.54                        & 35.44                        & {\color[HTML]{0000FF} 44.22} & {\color[HTML]{0000FF} 44.31} & {\color[HTML]{0000FF} 44.12} & {\color[HTML]{0000FF} 44.31} \\
BCR~\cite{an2024leveraging}                      &                        & {\color[HTML]{0000FF} 47.16} & {\color[HTML]{0000FF} 47.21} & {\color[HTML]{0000FF} 47.21} & {\color[HTML]{0000FF} 44.04} & {\color[HTML]{0000FF} 44.09} & {\color[HTML]{0000FF} 44.02} & 36.20                         & 36.57                        & 35.92                        & 36.57                        \\
PPR (Ours)              & \multirow{-5}{*}{5}    & \textbf{50.58}               & \textbf{50.63}               & \textbf{50.64}               & \textbf{47.34}               & \textbf{47.36}               & \textbf{47.31}               & \textbf{45.92}               & \textbf{46.06}               & \textbf{45.87}               & \textbf{45.88}               \\ \hline
ProtoNet+NLC~\cite{simon2022meta}             &                        & 39.71                        & 39.53                        & 39.59                        & 36.82                        & 36.84                        & 36.76                        & {\color[HTML]{0000FF} 45.73} & 45.66                        & {\color[HTML]{0000FF} 45.61} & {\color[HTML]{0000FF} 45.78} \\
RelationNet~\cite{simon2022meta}              &                        & {\color[HTML]{0000FF} 49.23}                       & {\color[HTML]{0000FF} 49.39}                      & {\color[HTML]{0000FF} 49.36}                       & 45.75                        & 45.84                        & 45.75                        & 40.17                        & 40.19                        & 40.08                        & 40.16                        \\
LPN+NLC~\cite{simon2022meta}                  &                        & 37.50                         & 37.53                        & 37.53                        & 36.93                        & 37.00                           & 36.92                        & 45.58                        & {\color[HTML]{0000FF} 45.91} & 45.32                        & 45.77                        \\
BCR~\cite{an2024leveraging}                      &                        & 47.90  &  47.86 &  47.99 & {\color[HTML]{0000FF} 46.31} & {\color[HTML]{0000FF} 46.34} & {\color[HTML]{0000FF} 46.31} & 43.58                        & 43.70                         & 43.61                        & 43.62                        \\
PPR (Ours)              & \multirow{-5}{*}{10}   & \textbf{50.63}               & \textbf{50.68}               & \textbf{50.64}               & \textbf{47.16}               & \textbf{47.19}               & \textbf{47.24}               & \textbf{48.49}               & \textbf{48.60}                & \textbf{48.45}               & \textbf{48.53}               \\ \hline
\end{tabular}
}
\end{table*}

\subsubsection{Individual disease performance comparison}
To further validate the effectiveness of detecting each novel disease, we assess the F1 score for all novel classes across three datasets, as detailed in Tables~\ref{tab:novel}. 
The results show that PPR achieves the highest F1 score and outperforms previous SoTA methods across all novel classes on all three datasets.
Although the data distribution and disease classes in CXR14~\cite{wang2017chestx} exhibit significant differences compared to CheXpert~\cite{irvin2019chexpert} and MIMIC~\cite{johnson2019mimic} datasets, PPR still achieves the best results among existing methods in both 5-shot and 10-shot settings.
While the CheXpert~\cite{irvin2019chexpert} and MIMIC~\cite{johnson2019mimic} datasets contain the same disease categories, they differ in their novel-class partitions due to differences in disease prevalence.
% The results further indicate that PPR maintains superior performance in both 5-shot and 10-shot settings. 
A slight reduction in F1 score from the 5-shot to the 10-shot setting on the MIMIC dataset, primarily due to higher recall accompanied by a minor drop in precision, which can also be observed in the overall performance comparison.
These results underscore that PPR consistently improves classification of low-prevalence diseases across all datasets, demonstrating robust, reliable performance even in the more challenging 5-shot scenario.

\begin{table}[]
\centering
\caption{Performance comparison results of cross-domain rare disease classification (\%) in the 3-way 5-shot setting. The best scores are highlighted in bold and the second-best in blue.}
\label{tab:cross}

\resizebox{\columnwidth}{!}{
% \resizebox{\textwidth}{!}{
% \footnotesize 
\large
\begin{tabular}{c|lccccccc}
\hline
Dataset                           & Method       & maF1                         & miF1                                  & maR                          & miR                          & maP                          & miP                                   & mAP                                   \\ \hline
                                  & ProtoNet+NLC~\cite{simon2022meta} & 45.67                        & {\color[HTML]{0000FF} 47.85}          & 47.89                        & 47.92                        & \textbf{47.64}               & \textbf{47.64}                        & 45.72                                 \\
                                  & RelationNet~\cite{simon2022meta}  & 39.07                        & 42.24                                 & 45.61                        & 45.67                        & 41.16                        & 41.22                                 & 54.32                                 \\
                                  & LPN+NLC~\cite{simon2022meta}      & 44.73                        & 47.07                                 & 47.38                        & 47.45                        & {\color[HTML]{0000FF} 47.08} & {\color[HTML]{0000FF} 47.13}          & 44.60                                  \\
                                  & BCR~\cite{an2024leveraging}          & {\color[HTML]{0000FF} 46.18} & 47.11                                 & {\color[HTML]{0000FF} 49.42} & {\color[HTML]{0000FF} 49.44} & 46.57                        & 46.56                                 & {\color[HTML]{0000FF} 56.41}          \\
\multirow{-5}{*}{CheXpert~\cite{irvin2019chexpert}→CXR-LT~\cite{holste2023cxr}} & PPR (Ours)  & \textbf{51.43}               & \textbf{51.70}                         & \textbf{62.61}               & \textbf{62.61}               & 44.28                        & 44.28                                 & \textbf{57.45}                        \\ \hline
                                  & ProtoNet+NLC~\cite{simon2022meta} & {\color[HTML]{0000FF}40.19}                       & 37.23                                 & 34.87                        & 36.92                        & 36.91 & 36.81 & 39.78                                 \\
                                  & RelationNet~\cite{simon2022meta}  & 40.12                        & 41.53                                 & {\color[HTML]{0000FF} 45.56} & {\color[HTML]{0000FF} 45.56} & 39.43                        & 39.43                                 & {\color[HTML]{0000FF} 49.78} \\
                                  & LPN+NLC~\cite{simon2022meta}      & 38.46                        & {\color[HTML]{0000FF} 42.58} & 40.92                        & 40.93                        & \textbf{44.20}                & \textbf{44.19}                        & 42.71                                 \\
                                  & BCR~\cite{an2024leveraging}          &  39.95 & 41.05                                 & 42.09                        & 42.03                        & {\color[HTML]{0000FF} 41.81}                        & {\color[HTML]{0000FF} 41.80}                                  & \textbf{52.95}                        \\
\multirow{-5}{*}{CXR14~\cite{wang2017chestx}→CXR-LT~\cite{holste2023cxr}}    & PPR (Ours)  & \textbf{42.66}               & \textbf{43.73}                        & \textbf{53.64}               & \textbf{53.70}                & 37.54                        & 37.59                                 & 48.50                                  \\ \hline
\end{tabular}

}
\end{table}

\subsubsection{Cross-domain rare disease evaluation}
Robust cross-domain generalization is particularly important for rare disease recognition, as collecting sufficient samples from every clinical institution is often impractical. 
However, substantial domain shifts may arise from differences in patient populations, disease prevalence, and imaging protocols.
% In real-world applications, there may be a domain shift between different datasets. 
To assess the cross-domain generalization capability of the proposed PPR framework for rare disease classes, we conduct experiments on the CXR-LT dataset~\cite{holste2023cxr} under a 3-way 5-shot setting.
As shown in Table.~\ref{tab:cross}, when trained on CheXpert~\cite{irvin2019chexpert} and evaluated on CXR-LT~\cite{holste2023cxr}, PPR achieves SoTA performance across nearly all metrics, particularly in F1 score and recall.
A similar trend is observed when training on CXR14~\cite{wang2017chestx} and testing on CXR-LT~\cite{holste2023cxr}, where PPR attains the highest F1 score and recall.
% Similarly, when trained on CXR14~\cite{wang2017chestx} and evaluated on CXR-LT~\cite{holste2023cxr}, PPR exhibits consistent trends, attaining the highest F1 score and recall.
These results confirm the strong generalization ability of PPR and its applicability for detecting rare diseases in real-world clinical settings.

\subsection{Ablation study}
\subsubsection{Ablation study of loss functions and prototype generation}
To evaluate the effectiveness of each design in PPR, we conduct a comprehensive ablation study across three datasets, as summarized in Table~\ref{tab:abl_loss}. 
% To validate the effectiveness of each design in PPR, we compare the different prototype generation methods as detailed in Table~\ref{tab:abl_gen}. 
The first three rows in each dataset compare different prototype generation strategies.
Specifically, “Baseline (mean)” denotes the simple averaging of support features~\cite{simon2022meta}, while “Baseline (ws)” represents a similarity-weighted sum aggregation~\cite{zhao2023few,an2024leveraging}.
Although “Baseline (ws)” slightly outperforms “Baseline (mean)” on the CheXpert dataset~\cite{irvin2019chexpert}, it exhibits inconsistent behavior across other datasets and suffers notable drops in F1 score and recall.
This observation suggests that biased reference features can propagate and substantially degrade the performance.
Compared to “Baseline (ws)”, the introduction of the proposed purification mechanism yields consistent and robust gains across all datasets, significantly improving macro- and micro-recall by 5.76\% and 5.77\% on CheXpert~\cite{irvin2019chexpert}, 22.91\% and 22.91\% on MIMIC~\cite{johnson2019mimic}, and 19.36\% and 19.35\% on CXR14~\cite{wang2017chestx}, respectively.
These results indicate the importance of purification in constructing prototypes for reliable multi-label classification in FSL.
Further incorporation of the proposed regulation strategy yields the best overall performance across all three datasets, with additional improvements in both F1 score and recall.
The overall results demonstrate that jointly considering prototype quality and disease relationships provides complementary benefits for representation learning and disease recognition in MLFSL.

\begin{table*}[t]
\centering
\caption{Ablation study of loss functions and different prototype generation methods in the 3-way 5-shot setting (\%).}
\label{tab:abl_loss}
\resizebox{\textwidth}{!}{

\begin{tabular}{c|lccccccc}
\hline
Dataset                   & Method                   & \multicolumn{1}{l}{maF1} & miF1                       & maR   & miR   & maP   & miP   & mAP   \\ \hline
\multirow{4}{*}{Chexpert~\cite{irvin2019chexpert}} & Baseline (mean)          & 41.62                     & \multicolumn{1}{l}{43.31} & 56.45 & 56.46 & 35.71 & 35.71 & \textbf{47.08} \\
& Baseline (ws)                               & 46.74                     & 47.66                     & 72.50                      & 72.51                     & \textbf{35.94}            & \textbf{35.94}            & 46.40  \\
                          
                          & Baseline (ws)+ $\mathcal{L}_{\text{pur}}$           & 48.27                     & 48.98                      & 78.26 & 78.28 & 35.77 & 35.78 & 46.93 \\
                          
                          & PPR                       & \textbf{49.82}                     & \textbf{50.41}                      & \textbf{86.84} & \textbf{86.85} & 35.72 & 35.72 & 46.37 \\ \hline
\multirow{4}{*}{MIMIC~\cite{johnson2019mimic}}& Baseline (mean)          & 43.08                     & 44.04                      & 58.83 & 58.83 & \textbf{35.79} & \textbf{35.79} & \textbf{46.94} \\    & Baseline (ws)          & 38.18                     & 38.94                     & 45.15                     & 45.15                     & 35.38                     & 35.37                     & 46.42 \\
                          
                          & Baseline (ws) + $\mathcal{L}_{\text{pur}}$          & 45.53                     & 46.45                      & 68.06 & 68.06 & 35.43 & 35.44 & 46.66 \\
                          
                          & PPR                       & \textbf{46.27}                     & \textbf{47.08}                      & \textbf{70.87} & \textbf{70.88} & 35.54 & 35.55 & 46.67 \\ \hline
\multirow{4}{*}{CXR14~\cite{wang2017chestx}}& Baseline (mean)          & 38.88                     & 40.88                      & 44.87 & 44.87 & 39.85 & 39.84 & \textbf{51.89} \\    & Baseline (ws)          & 31.69                     & 34.36                     & 30.61                     & 30.62                     & \textbf{44.01}                     & \textbf{43.96}            & 51.85 \\
                          
                          & Baseline (ws) + $\mathcal{L}_{\text{pur}}$         & 42.59                     & 43.77                      & 49.97 & 49.97 & 40.11 & 40.12 & 51.51 \\
                          
                          & PPR                       & \textbf{43.93}                     & \textbf{44.92}                      & \textbf{53.78} & \textbf{53.78} & 40.12 & 40.10  & 51.64 \\ \hline
\end{tabular}

}
\end{table*}

\begin{table*}[]
\centering
\caption{Ablation study of different prototype regulation methods in the 3-way 5-shot setting (\%).}
\label{tab:abl_reg}
\resizebox{\textwidth}{!}{

\begin{tabular}{c|lccccccc}
\hline
Dataset                   & Method                   & \multicolumn{1}{l}{maF1} & miF1                       & maR   & miR   & maP   & miP   & mAP   \\ \hline
\multirow{3}{*}{Chexpert~\cite{irvin2019chexpert}} 
            
                          & max                      & 47.09                     & 48.47                      & 76.13 & 76.14 & \textbf{35.83} & \textbf{35.83} & 46.55 \\
                          & min                      & 47.25                     & 48.24                      & 75.26 & 75.27 & 35.78 & 35.78 & \textbf{46.69} \\
                          & PPR                       & \textbf{49.82}                     & \textbf{50.41}                      & \textbf{86.84} & \textbf{86.85} & 35.72 & 35.72 & 46.37 \\ \hline
\multirow{3}{*}{MIMIC~\cite{johnson2019mimic}}    
                          & max                      & 40.51                     & 42.6                       & 55.53 & 55.53 & 35.52 & 35.51 & \textbf{46.82} \\
                          & min                      & 42.54                     & 43.66                      & 58.54 & 58.54 & \textbf{35.74} & \textbf{35.73} & 46.49 \\
                          & PPR                       & \textbf{46.27}                     & \textbf{47.08}                      & \textbf{70.87} & \textbf{70.88} & 35.54 & 35.55 & 46.67 \\ \hline
\multirow{3}{*}{CXR14~\cite{wang2017chestx}}    
                          & max                      & 38.64                     & 39.97                      & 40.76 & 40.76 & \textbf{40.48} & \textbf{40.46} & 51.54 \\
                          & min                      & 42.59                     & 43.77                      & 49.97 & 49.97 & 40.11 & 40.12 & 51.51 \\
                          & PPR                       & \textbf{43.93}                     & \textbf{44.92}                      & \textbf{53.78} & \textbf{53.78} & 40.12 & 40.10  & \textbf{51.64} \\ \hline
\end{tabular}

}
\end{table*}

\subsubsection{Ablation study of prototype regulation}
To validate our adaptive regulation strategy, we further compare different inter-class separation strategies, as summarized in Table~\ref{tab:abl_reg}.
The term “max” denotes the conventional approach that enforces maximal inter-class separation in the feature space~\cite{hersche2022constrained,yang2022few}, whereas “min” follows the CLOSER~\cite{oh2024closer} assumption, encouraging reduced inter-class distance to exploit shared information.
The comparison results of different separation strategies demonstrate that our proposed adaptive regulation method, which regulates inter-class similarity based on disease comorbidity statistics, consistently achieves the best performance across all datasets.
These results highlight the importance of considering the intricate relationship between diseases in MIA, as shared pathological characteristics and comorbidity correlations can significantly impact MLFSL performance.

\begin{figure*}[t]
  \centering
  \includegraphics[width=\textwidth]
  {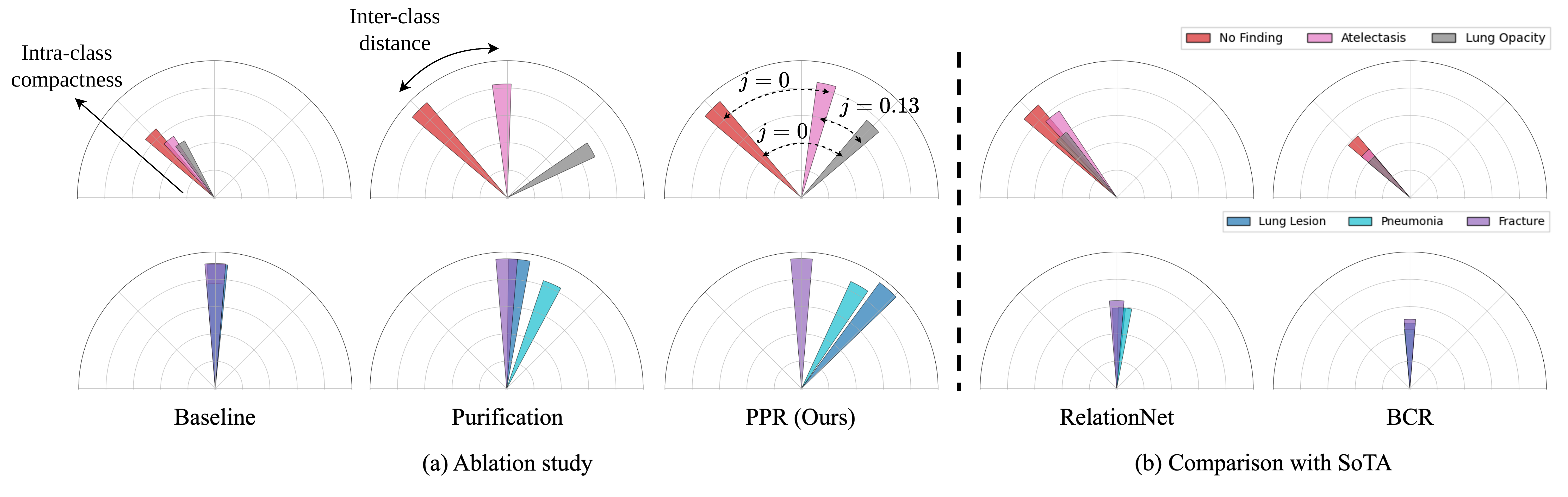} %vis2
  \caption{Visualization of representations trained on CheXpert in the 3-way 5-shot setting. 
  The radius indicates intra-class compactness, with bars closer to the circumference denoting higher intra-class feature similarity. 
  The angle reflects inter-class distance, where a wider angle corresponds to a greater distance between disease classes.
  The dotted line denotes the comorbidity statistics $j$ between two diseases used in training.
  Base and novel classes are displayed in the upper and lower plots, respectively.
  % The upper plot illustrates the base classes, while the lower plot presents the novel classes. 
  % The purification mechanism enhances intra-class feature compactness and inter-class discriminability.
  % Moreover, introducing the regulation strategy (PPR) further organizes the feature representations according to disease relationships, forming a clinically relevant embedding space.
  % For instance, \textit{Lung Opacity} and \textit{Atelectasis} show higher comorbidity scores than their pairs with \textit{No Finding} in the base class set.
  % In the novel class set, \textit{Pneumonia} and \textit{Lung Lesion} exhibit stronger clinical correlation than \textit{Fracture}, which shows limited association with other diseases~\cite{irvin2019chexpert}.
  }
  \label{fig:vis}
\end{figure*}

% Notably, the consistent improvement in macro-level metrics demonstrates that PPR could improve overall classification ability for novel class diseases, which can be extension for rare diseases at the same time, addressing a critical issue in MIA. 

% with the macro-F1, micro-F1, macro-recall, and micro-recall enhanced by 1.55\%, 1.43\%, 8.58\%, and 8.57\% on CheXpert and 0.74\%, 0.63\%, 2.81\%, and 2.82\% on MIMIC.

\subsubsection{Visualization}
To intuitively illustrate the effect of the proposed purification and regulation mechanisms, we visualize the learned embedding spaces in Fig.~\ref{fig:vis}.
In the radial visualization, the radius measures intra-class compactness, while the angular distance between bars indicates inter-class separation.
% In the radial visualization, the radius represents intra-class compactness, where bars closer to the circumference indicate higher feature similarity within each class. 
% The angular distance between bars reflects inter-class separation, with wider angles indicating greater semantic distances between diseases.
As demonstrated in Fig.~\ref{fig:vis}a, the purification strategy promotes compact intra-class clustering and effectively disentangles mixed feature distribution caused by multi-label co-occurrence.
Moreover, incorporating the regulation method (PPR) further structures the embedding space according to comorbidity correlation, where diseases with higher co-occurrence frequencies are positioned closer together than those with lower co-occurrence diseases. 
% On the other hand, the comparison with other SoTA methods is illustrated in Fig.~\ref{fig:vis}b.
On the other hand, when comparing with other SoTA methods (Fig.~\ref{fig:vis}b), although RelationNet~\cite{simon2022meta} and BCR~\cite{an2024leveraging} achieve comparable quantitative performance, their learned representations remain entangled within the embedding space, thereby compromising inter-class discriminability.
These results confirm that PPR effectively disentangles co-occurring class features while accounting for the inter-disease relationships that are overlooked by prior studies.

\subsubsection{Hyperparameter analysis of $\delta$}
\begin{figure*}[t]
  \centering
  \includegraphics[width=\textwidth]
  {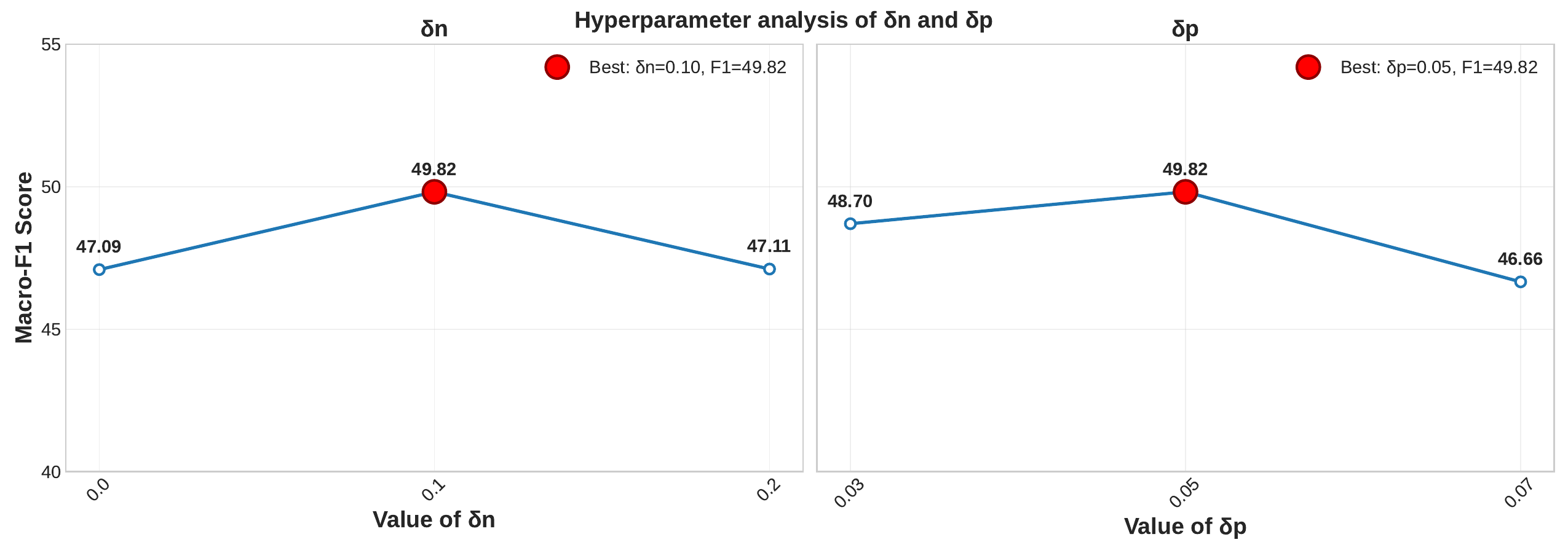} %vis2
  \caption{Hyperparameter analysis of $\delta$ on CheXpert in the 3-way 5-shot setting. 
  }
  \label{fig:del}
\end{figure*}

To evaluate the robustness of the proposed regulation strategy, we analyze the sensitivity of the margin hyperparameters $\delta_p$ and $\delta_n$, as shown in Fig.~\ref{fig:del}.
The best F1 score is achieved with $\delta_n=0.1$, indicating that introducing an appropriate negative margin effectively enlarges the inter-class semantic separation and improves class discrimination. 
Meanwhile, $\delta_p=0.05$ yields the highest F1 score by allowing a mild similarity tolerance between clinically related diseases, thereby preserving feature sharing while preventing excessive prototype separation. 
These results demonstrate that the proposed regulation strategy is robust to margin selection and that properly balancing prototype compactness and inter-class separability is beneficial for MLFSL.

\subsubsection{Hyperparameter analysis of $\lambda$}
\begin{figure*}[]
  \centering
  \includegraphics[width=\textwidth]
  {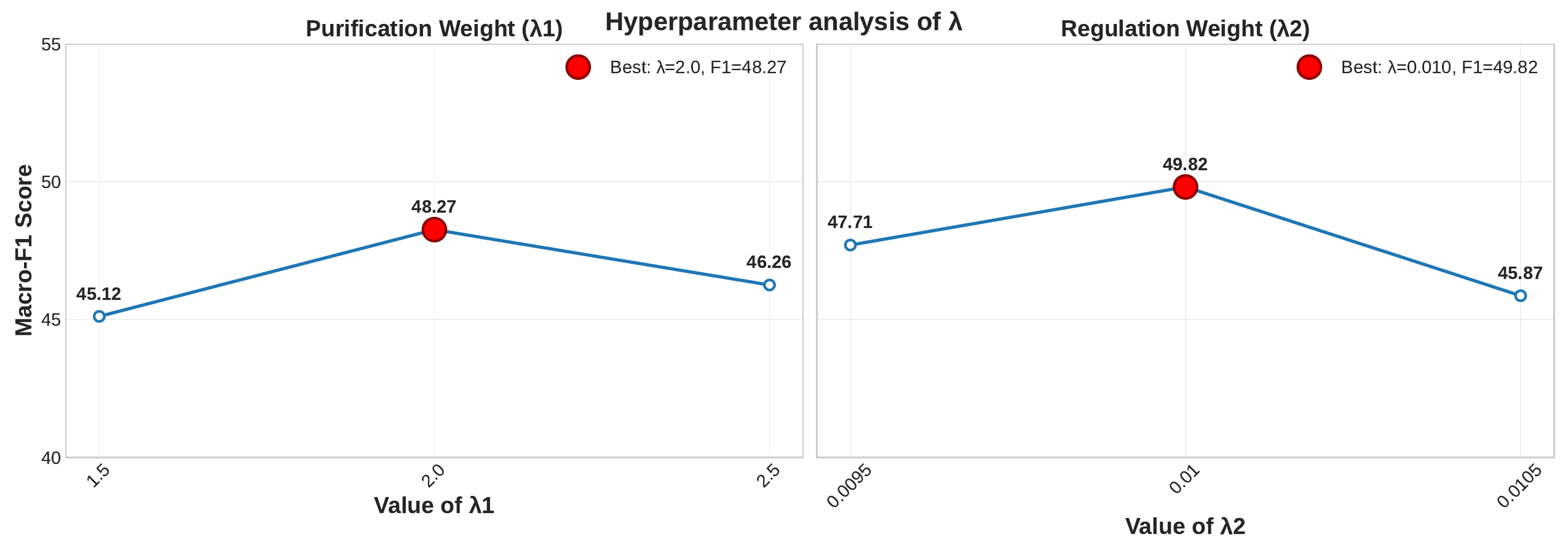} %vis2
  \caption{Hyperparameter analysis of $\lambda$ on CheXpert in the 3-way 5-shot setting. 
  }
  \label{fig:lam}
\end{figure*}

To evaluate the sensitivity of the proposed framework to the loss-balancing coefficients, we perform a hyperparameter analysis on the CheXpert dataset under the 3-way 5-shot setting, as shown in Fig.~\ref{fig:lam}.
The coefficients $\lambda_1$ and $\lambda_2$ control the relative contributions of the purification and regulation losses, respectively.
Since prototype purification is performed prior to prototype regulation, we first analyze the effect of $\lambda_1$ and subsequently evaluate $\lambda_2$ using the selected $\lambda_1$.
A grid search is conducted on each dataset, and the final coefficients are determined based on the F1 score and applied consistently throughout all experiments.
When $\lambda_2$ is too small, the regulation objective is underweighted, limiting its ability to capture clinically meaningful disease relationships.
Conversely, excessively large $\lambda_2$ values cause the regulation objective to dominate optimization, reducing the complementary effect of prototype purification.
Overall, these results demonstrate that PPR is robust to the selection of the loss-balancing coefficients, while appropriately balancing the purification and regulation objectives consistently yields superior performance.

\section{Discussion}
% experiment conclusion
% limitation and future work
% Overall conclusion

MLFSL remains a challenging problem in MIA because disease annotations are often scarce and multiple diseases may coexist within a single image. 
Under such settings, metric-based approaches have attracted considerable attention due to their efficiency, interpretability, and strong performance in MIA~\cite{quinonez2025comparative,pachetti2024systematic}.
However, constructing reliable prototypes becomes difficult when disease features are entangled by comorbidity patterns, potentially leading to biased disease representations and degraded recognition performance.
To address these challenges, we propose PPR, a novel MLFSL framework that improves representation learning through prototype purification and comorbidity-aware regulation, thereby enhancing disease detection performance.
Furthermore, we adopt a more realistic clinical evaluation protocol in which base and novel diseases are divided by disease prevalence and remain mutually exclusive between the training and test sets, better reflecting real-world scenarios in which models must generalize to underrepresented diseases.

The experimental results suggest that prototype quality plays a critical role in MLFSL. 
By introducing prototype purification, PPR rectifies the reference feature used for similarity-weighted aggregation and generates cleaner prototypes by emphasizing disease-representative features during prototype construction, thereby reducing the influence of irrelevant disease information. 
The ablation studies demonstrate that purified prototypes consistently improve disease detection performance across datasets, highlighting the importance of mitigating prototype contamination in MLFSL.

Beyond prototype construction, our findings indicate that disease relationships should be considered during representation learning. 
Previous studies have reported conflicting observations regarding inter-class separation~\cite{li2020boosting,hersche2022constrained,yang2023neural,song2023learning,jiang2022multi}, with some advocating larger class margins for improved discrimination and others emphasizing feature similarity for better novel-class transferability from base class~\cite{oh2024closer}. 
These observations imply that a uniform separation strategy may not be suitable for all categories. 
In MIA, diseases frequently exhibit clinically meaningful comorbidity patterns and shared visual characteristics. 
Motivated by this, the proposed prototype regulation strategy incorporates comorbidity statistics to adaptively regulate prototype similarity. 
This design allows the embedding space to better reflect clinically meaningful disease associations while preserving class discriminability, resulting in improved overall performance.

An important advantage of the proposed framework is its simplicity. 
Unlike previous methods that require additional modalities~\cite{fang2025inferring,li2026attribute} or auxiliary modules~\cite {simon2022meta,an2024leveraging}, PPR operates directly on feature representations and disease labels. 
This design introduces limited computational overhead while remaining compatible with existing metric-based few-shot learning frameworks. 
Such flexibility may facilitate its adoption in practical MIA applications.

Another important finding is the strong cross-domain performance achieved by PPR. 
The consistent improvements observed in cross-domain evaluation suggest that the proposed framework captures transferable disease representations and may therefore be useful in real-world clinical scenarios where annotated samples are limited.
Generalization to unseen datasets remains a major challenge in MIA due to variations in disease prevalence, patient populations, and imaging protocols. 
The consistent performance improvements obtained under cross-domain settings suggest that PPR learns transferable disease representations rather than dataset-specific patterns. 
This property is particularly important for rare disease recognition, where collecting large-scale annotated datasets is often impractical.

% limitation
\noindent {\textbf{Limitations and future works.}} Despite these promising results, several limitations should be acknowledged. 
First, the proposed framework estimates relative feature purity prior using label-derived information and assumes that samples with fewer co-occurring diseases provide relatively purer disease representations. 
Although this assumption is clinically intuitive and supported by the experimental results, disease labels may not fully capture the complexity of disease manifestations in medical images.
Consequently, the estimated purity prior may not fully reflect the underlying feature quality. 
Future research will investigate more sophisticated purity estimation strategies that jointly consider feature representations and label information to better characterize disease-specific features.

Second, the prototype regulation strategy relies on disease comorbidity statistics derived from the training data to model inter-class relationships. 
While comorbidity provides clinically meaningful information, disease prevalence and comorbidity patterns may vary across institutions, patient populations, and clinical settings.
Such distribution shifts could affect the reliability of the estimated disease associations.
Future work will explore more adaptive approaches to account for the disease relationships to improve robustness across diverse clinical environments.

Finally, the proposed framework is designed for image-level disease classification. 
Although the proposed purification and regulation mechanisms demonstrate effectiveness in MLFSL, their applicability to other MIA tasks remains unexplored.
Future research may extend the proposed framework to other tasks, such as lesion localization, segmentation, and prognosis prediction, thereby broadening its clinical applicability.

\section{Conclusion}
In this work, a novel MLFSL framework for MIA, PPR, is proposed to address prototype contamination and the inter-class feature separation problem.
Comprehensive experiments across four benchmark datasets, including cross-domain evaluation, demonstrate that PPR consistently achieves SoTA performance, underscoring its effectiveness and generalizability for clinical practice.
A future direction is to validate our framework on additional medical image modalities to improve generalization.
Moreover, extending PPR to multi-modal medical data presents a promising direction for constructing more purified prototypes.

\bibliographystyle{splncs04}
\bibliography{main}


% %% For numbered reference style

%   \textit{\LaTeX: a document preparation system},
%   Addison Wesley, Massachusetts,

% \end{thebibliography}

\end{document}